\documentclass[journal]{IEEEtran}

\ifCLASSINFOpdf
\else
   \usepackage[dvips]{graphicx}
\fi
\usepackage{url}

\usepackage{graphicx}
\usepackage{subcaption}

\usepackage{xcolor}
\usepackage{cite}
\usepackage{amssymb,amsfonts, amsmath, amsthm}

\usepackage{algorithm,algpseudocode}
\algrenewcommand\algorithmicindent{0.7em}

\usepackage{physics}

\begin{document}

\title{A Bayesian Filtering Approach for Learning Lagrangian Dynamics from Noisy Measurements
}

\author{Kundan Kumar$^{*}$,  Shreya Das$^{*}$, and Simo Särkkä, \IEEEmembership{Senior Member, IEEE}
 \thanks{$^{*}$Equal contribution.}
 \thanks{K. Kumar is with the School of Engineering and Applied Science, Ahmedabad University, India (e-mail: kundan.kumar@ahduni.edu.in).
 
 S. Das is with the Department of Electrical Engineering, Indian Institute of Technology (BHU) Varanasi, India (e-mail: shreyadas.eee@iitbhu.ac.in).
 
 K. Kumar and S. Särkkä are with the ELLIS Institute Finland and the Department of Electrical Engineering and Automation, Aalto University, Finland (e-mail: \{kundan.kumar; simo.sarkka\}@aalto.fi).}
}

\maketitle

\begin{abstract}
This paper proposes a Bayesian filtering-based approach for learning the dynamics of a physical system from partial, noisy measurements. We model the system dynamics using a Lagrangian mechanics formulation. As in Lagrangian neural networks (LNNs), we parameterize the kinetic and potential energies with neural networks. The unknown external forces in the Lagrangian formulation are modeled as white Gaussian noise. The corresponding Euler--Lagrange equations then yield a continuous-time stochastic state-space model (SSM) that describes the system dynamics. The neural network parameters and system states are then jointly learned via a maximum-likelihood method using Gaussian-approximation-based Bayesian filters. The effectiveness of the proposed method is demonstrated on pendulum and Duffing oscillator examples, and its performance is compared with conventional LNNs and with approximate Bayesian filters using known system models.
\end{abstract}


\IEEEpeerreviewmaketitle

\section{Introduction}
Accurate modeling of the dynamics of a physical system is crucial in many applications, including robotics \cite{siciliano2008springer}, industrial automation \cite{rigatos2011modelling}, navigation \cite{grewal2007global}, health technology \cite{sarkka2012dynamic, sarkka2019applied}, and target tracking \cite{li2003survey}. In many applications, constructing the model from the first principles is not possible because the physics are either partially or fully unknown. In those cases, it is possible to treat the system unknown and learn the dynamic model from data using approaches such as neural networks (NNs) \cite{kumpati1990identification, chu2002neural}, basis function expansions \cite{svensson2015nonlinear}, or Gaussian process regression \cite{svensson2016computationally, sarkka2019use}. However, purely data-driven approaches typically require large amounts of training data and often fail to obey the known physical laws. Physics-informed neural networks (PINNs) address these limitations by combining data-driven learning with physical constraints, thereby improving generalization and interpretability while reducing data requirements \cite{karniadakis2021physics, cai2021physics, greydanus2019hamiltonian}. 

\begin{figure}
    \centering
    \includegraphics[width=8cm, height=4.5cm]{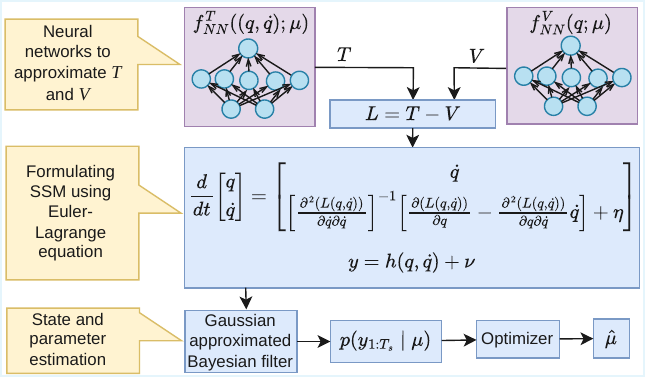}
    \caption{Schematic diagram of the proposed method. The kinetic energy ($T$) and potential energy ($V$) of the Lagrangian ($L$) are parameterized using neural networks (NNs). The Euler--Lagrange equation with stochastic input is rewritten as a state-space model. The state and the neural network parameters are jointly estimated with an approximate Bayesian filter.} 
    \label{fig_main_intro}
\end{figure}

Lagrangian neural networks (LNNs) are a class of PINNs that use a Lagrangian formulation to learn system dynamics \cite{cranmer2020lagrangian}.  By leveraging the Lagrangian formulation, LNNs incorporate physical laws, such as energy conservation, directly into the learning process. Deep Lagrangian neural networks (DeLaN) were introduced in \cite{lutterdeep} as a physics-informed approach for learning the equations of motion of robotic systems in real time. Subsequent work extended this framework to model system energy and friction via inverse learning and demonstrated its effectiveness in real-world robotic tasks and reinforcement learning \cite{lutter2019energy, lutter2021combining, das2026integrating}. 

Existing formulations of LNN typically assume access to the full system states, including the generalized velocities, which is often unrealistic in real-world settings. When only generalized coordinates are measured, the generalized velocities can, in principle, be obtained via numerical differentiation \cite{atkinson2008introduction}. However, this approach is limited in generality, amplifies measurement noise, and can significantly degrade performance. In this paper, we propose an alternative approach, where an approximate Bayesian filter, such as an extended Kalman filter (EKF) \cite{bar2001estimation,sarkka2023bayesian} or a sigma-point filter \cite{arasaratnam2009cubature, wan2000unscented, ito2002gaussian} is used to estimate the full system states from partial and noisy measurements jointly with the LNN learning task. In this approach, the neural networks used to parameterize the Lagrangian are trained via a parameter estimation procedure implemented on top of the filter, so the training explicitly accounts for measurement partiality and noise.

Specifically, the main contributions of this paper are as follows (see also Fig.~\ref{fig_main_intro}):
\begin{itemize}
    \item We formulate the Lagrangian dynamics learning problem from partial and noisy measurements within a Bayesian state estimation framework.

    \item We model the unknown forces and uncertainties entering the Euler--Lagrange equations as white Gaussian noises, which leads to a continuous-time stochastic state space model (SSM). 

    \item The kinetic and potential energies of the Lagrangian are parameterized using neural networks and embedded into the Euler--Lagrange equation.
    
    \item We employ an approximate Bayesian filtering method to jointly estimate the system states and learn the neural network parameters from partial observations.
\end{itemize}
Unlike conventional LNNs that rely on full-state observations, the proposed approach uses noisy partial observations. Its effectiveness is demonstrated through numerical experiments, and its performance is compared with that of traditional LNNs and Gaussian-approximated Bayesian filters using the known model.

\section{Lagrangian Formulation}
Lagrangian mechanics is a way to formulate the dynamics of a physical system by expressing them in terms of the system's kinetic and potential energies \cite{goldstein1950classical, lee2018global}. Consider a system with generalized coordinate $q=(q_1,q_2,\ldots,q_n)$ and the corresponding generalized velocity given by $\dot{q}=\frac{dq}{dt}$. Let us assume that the potential energy is a scalar function $V(q)$ defined on the configuration space and depends only on the generalized coordinates $q$. The kinetic energy $T(q, \dot{q})$ can be a function of both the generalized coordinate $q$ and the velocity $\dot{q}$, although it often only depends on $\dot{q}$. The Lagrangian can then be written as follows \cite[p.~35]{goldstein1950classical}:
\begin{equation}\label{Lagrange_eqn}
  L(q,\dot{q}) = T(q,\dot{q}) - V(q).
\end{equation}
The Euler--Lagrange equations for the corresponding system are given by \cite[p.~48]{goldstein1950classical}
\begin{equation}\label{EL_eqn}
   \frac{d}{dt} \frac{\partial L(q, \, \dot{q})}{\partial \dot{q}} - \frac{\partial L(q, \, \dot{q})}{\partial q} = \eta_\tau,
\end{equation}
where term $\eta_\tau$ accounts for the non-conservative effects, including external forces. By applying the chain rule, \eqref{EL_eqn} becomes 
\begin{equation}\label{Eq:EL_eqn_chain}
 \frac{\partial^2 L(q,\dot{q})}{\partial q \partial \dot{q}}\dot{q}+\frac{\partial^2 L(q,\dot{q})}{\partial \dot{q} \partial \dot{q}}\ddot{q} - \frac{\partial L(q, \, \dot{q})}{\partial q} = \eta_\tau.
\end{equation}
We assume that the Hessian matrix, $\frac{\partial^2 L(q,\dot{q})}{\partial \dot{q}\partial \dot{q}}$, is non-singular. The above equation \eqref{Eq:EL_eqn_chain} can then be explicitly solved for the generalized acceleration, $\ddot{q}$, as follows:
\begin{equation}\label{eq_qddot_formula}
    \ddot{q} = \Big[ \frac{\partial^2 L(q, \, \dot{q})}{\partial \dot{q} \partial \dot{q}} \Big]^{-1} \, \Big[ \eta_\tau+\frac{\partial L(q, \, \dot{q}) }{\partial q} - \frac{\partial^2 L(q, \, \dot{q})}{\partial q \partial \dot{q}} \dot{q}   \Big].
\end{equation}
We can now model the following term as white noise:
\begin{equation}\label{Eq:white_noise}
\eta \approx \Big[ \frac{\partial^2 L(q, \, \dot{q})}{\partial \dot{q} \partial \dot{q}} \Big]^{-1}\eta_\tau,
\end{equation}
where $\eta$ is a Gaussian white noise with a spectral density $Q_c$. Using \eqref{eq_qddot_formula} and \eqref{Eq:white_noise}, the continuous-time dynamics of a system can be expressed in first-order stochastic state space form in terms of the state vector $\begin{bmatrix}
    q & \dot{q}
\end{bmatrix}^\top$ as follows: 
\begin{equation}\label{Eq_process_LNN1}
\begin{split}
  &  \frac{d}{dt} \begin{bmatrix}
       q \\
        \dot{q}
    \end{bmatrix}  = \begin{bmatrix}
        \dot{q}\\
        \Big[ \frac{\partial^2 \left(L(q,\dot{q})\right)}{\partial \dot{q} \partial \dot{q}} \Big]^{-1} \, \Big[ \frac{\partial \left(L(q,\dot{q})\right) }{\partial q} - \frac{\partial^2 \left(L(q, \dot{q})\right) }{\partial q \partial \dot{q}} \dot{q}   \Big]
    \end{bmatrix}
    +
    \begin{bmatrix}
      0 \\ \eta
    \end{bmatrix}.
\end{split}
\end{equation}

%

\section{Proposed Methodology}
In this section, we develop a Bayesian estimation framework that jointly learns the system dynamics and estimates the system states from partial and noisy measurements within the Lagrangian formulation.

\subsection{LNN learning and Bayesian filtering}

Let us assume that the kinetic and potential energies, $T(q, \, \dot{q})$ and $V(q)$, are not available a priori and hence the Lagrangian $L(q, \dot{q})$ of the system is unknown. To learn the system dynamics, our aim is now to infer the unknown energy functions from the available measurements using a suitable class of function approximators. In this work, we employ a neural network-based approach to approximate the energy functions. This is the same approach as is used in Lagrangian neural networks (LNNs) \cite{cranmer2020lagrangian,lutterdeep,lutter2019energy,lutter2021combining}.

The unknown $T(q, \dot{q})$ and $V(q)$ are approximated using neural networks as follows:
\begin{equation}\label{Eq_approx_T_V_NN}
    \begin{split}
        T(q, \dot{q}) &\approx T_{\text{NN}}(q, \,\dot{q}; \mu), \\
        V(q) &\approx V_{\text{NN}}(q; \mu), 
    \end{split}
\end{equation}
where $T_{\text{NN}}$ and $V_{\text{NN}}$ denote neural networks (NNs), and $\mu$ represents the learnable parameters (weights and biases). The acceleration term in \eqref{Eq_process_LNN1} can then be approximated as:
\begin{equation}\label{Eq_NN-appr_accln}
    \ddot{q} \approx f(q, \, \dot{q}; \, \mu)+\eta,
\end{equation}
where
\begin{equation*}
\begin{split}
  L_{\text{NN}}(q, \,\dot{q}; \mu) &= T_{\text{NN}}(q, \,\dot{q}; \mu) - V_{\text{NN}}(q; \mu), \\
  f(q, \, \dot{q}; \, \mu) &= \Bigg[ \frac{\partial^2 \left(L_{\text{NN}}(q,\dot{q}; \mu)\right)}{\partial \dot{q} \partial \dot{q}} \Bigg]^{-1} \\
  &\quad \times \Bigg[ \frac{\partial \left(L_{\text{NN}}(q,\dot{q}; \mu)\right) }{\partial q} - \frac{\partial^2 \left(L_{\text{NN}}(q, \dot{q}; \mu)\right) }{\partial q \partial \dot{q}} \dot{q} \Bigg].
\end{split}
\end{equation*}
In state estimation \cite{sarkka2023bayesian}, the state of the dynamical system is typically denoted by $x$. Defining the state as $ x = \begin{bmatrix}
    x_1 & x_2
\end{bmatrix}^\top  = \begin{bmatrix}
    q & \dot{q}
\end{bmatrix}^\top \in \mathbb{R}^{n_x}$, \eqref{Eq_process_LNN1} can be rewritten as
\begin{equation}\label{Eq_process_continuous}
    \begin{bmatrix}
        \frac{d x_1}{d t} \\
        \frac{d x_2}{d t}
    \end{bmatrix} = \begin{bmatrix}
        x_2 \\
        f(x_1, x_2; \mu)
    \end{bmatrix} + \begin{bmatrix}
        0 \\ \eta
    \end{bmatrix}.
\end{equation}
To enable state and parameter estimation using Bayesian filtering, the continuous-time dynamics can be discretized using, for example, the Euler--Maruyama method \cite{sarkka2019applied, sarkka2023bayesian}. The resulting discrete-time process model then has the form
\begin{equation*}
    \begin{bmatrix}
        x_{1, k} \\
        x_{2, k}
    \end{bmatrix} = \underbrace{\begin{bmatrix}
        x_{1, k-1} + x_{2, k-1} \Delta t \\
       x_{2, k-1}  + f(x_{1, k-1}, x_{2, k-1}; \mu)  \Delta t
    \end{bmatrix}}_{\tilde{f}(x_{k-1};\mu)} + \tilde{\eta}_{k-1},
\end{equation*}
where $\Delta t$ is the sampling interval and $\tilde{\eta}_{k-1} \sim \mathcal{N}(0, Q_{k-1})$ is a Gaussian process noise with $Q_{k-1} = \mathrm{diag}(0, Q_c \Delta t)$, which thus has the form
\begin{equation}\label{Eq_dynamic_model}
    x_k = \tilde{f}(x_{k-1};\mu) + \tilde{\eta}_{k-1}.
\end{equation}
Note that in practice, it is often beneficial to use more sophisticated discretization methods than the plain Euler--Maruyama (see, e.g, \cite{sarkka2019applied, sarkka2023bayesian}).

We further assume that the available data can be modeled as the following measurement model:
\begin{equation}\label{Eq_meas_model}
    y_k = h(x_k) + \nu_k,
\end{equation}
where $\nu_k \sim \mathcal{N}(0, R_k)$ is the Gaussian measurement noise. For example, if we only measure the generalized coordinates (but not the velocities), we have $h(x_k) = x_{1,k}$, but substantially more general measurement models can also be used.

Equations~\eqref{Eq_dynamic_model} and \eqref{Eq_meas_model} now define a standard Bayesian state estimation problem \cite{sarkka2023bayesian}, where the unknown parameters of the system are the neural network weights $\mu$. Thus, this formulation enables the joint state and parameter estimation within the Bayesian filtering framework.
\subsection{Affine approximation}\label{subsec_affine_approx}
The state and parameter estimation problem for the nonlinear state space model (SSM) \eqref{Eq_dynamic_model}-\eqref{Eq_meas_model} cannot be solved exactly in closed form. Therefore, we adopt a Gaussian approximated Bayesian filtering framework \cite{sarkka2023bayesian, lefebvre2002comment, garcia2015posterior, tronarp2018iterative} and construct an affine approximation of the stochastic SSM \eqref{Eq_dynamic_model}-\eqref{Eq_meas_model} as follows: 
\begin{equation}\label{Eq_affine_approx}
    \begin{split}
        	x_{k} & \approx A_{k-1} x_{k-1} + a_{k-1} + \acute{\eta}_{k-1}, \\
	y_k & \approx H_k x_k + b_k + \acute{\nu}_k, 
    \end{split}
	\end{equation}
where $\acute{\eta}_{k-1} \sim \mathcal{N}(0, \Lambda_{k-1})$ and  $\acute{\nu}_{k} \sim \mathcal{N}(0, \Omega_{k})$. The approximation parameters are $A_{k-1} \in \mathbb{R}^{n_x \times n_x}, a_{k-1} \in \mathbb{R}^{n_x}, H_{k} \in \mathbb{R}^{n_{y} \times n_x}$, and  $b_k \in \mathbb{R}^{n_y}$. Various approaches exist in the literature for computing these parameters. In this work, we employ 1) sigma-point-based statistical linear regression (SLR) and 2) the first-order Taylor series approximation (the extended Kalman filter, EKF). These approaches are discussed in detail below.

1) In statistical linear regression, the approximation parameters $A_{k-1},\;a_{k-1},\;\Lambda_{k-1}$ are computed as \cite{sarkka2023bayesian, garcia2015posterior, arasaratnam2007discrete} 
\begin{align*}
    A_{k-1}&= \Gamma^\top P_{k-1\mid k-1}^{-1},\\
    a_{k-1}&= \bar{x} -A_{k-1}\hat{x}_{k-1\mid k-1},\\
    \Lambda_{k-1}&=\Phi -A_{k-1}P_{k-1\mid k-1}A_{k-1}^\top,
\end{align*}
where $\hat{x}_{k-1\mid k-1} \approx E[x_{k-1}\mid y_{1:k-1}]$ is the approximate conditional mean of $x_{k-1}$ given measurements up to time step $k-1$, and the corresponding error covariance $P_{k-1\mid k-1} \approx E[(x_{k-1} - \hat{x}_{k-1 \mid k-1})(x_{k-1} - \hat{x}_{k-1 \mid k-1})^\top \mid y_{1:k-1}]$. Above, $\bar{x} \approx \sum_{j=1}^m \omega_j\tilde{f}(\zeta_{j, k-1\mid k-1}; \mu)$, $\Gamma = \sum_{j=1}^mw_j(\zeta_{j, k-1\mid k-1} - \hat{x}_{k-1\mid k-1})(\mathcal{Z}_j- \bar{x})^\top$, $\Phi = \sum_{j=1}^mw_j(\mathcal{Z}_j-\bar{x})(\mathcal{Z}_j-\bar{x})^\top$, 
where $\zeta_{j, k-1\mid k-1} = S_{k-1\mid k-1} \xi_j +\hat{x}_{k-1\mid k-1}$, $S_{k-1\mid k-1}$ is the lower triangular Cholesky factor of $P_{k-1\mid k-1}$, $m$ denotes the number of sigma points, $w_j$ is the weight corresponding to $j$th unit sigma point $\xi_j$, and $\mathcal{Z}_j=\tilde{f}(\zeta_{j, k-1\mid k-1}; \mu)$. The approximation parameters for the measurement model $(H_k, b_k, \Omega_k)$ are computed in a similar manner (see \cite{sarkka2023bayesian}).

2) In the Taylor series approximation, first-order Taylor series approximations \cite{bar2001estimation} of the nonlinear functions $\tilde{f}(x_{k-1})$ and $h(x_k)$ are employed around the available $\hat{x}_{k-1\mid k-1}$ and $\hat{x}_{k\mid k-1}$, respectively. Here, $\hat{x}_{k\mid k-1} \approx E[x_k\mid y_{1:k-1}]$ is the approximate mean of $x_k$ given $y_{1:k-1}$. The resulting approximation parameters are given by $A_{k-1} = F_j(\hat{x}_{k-1\mid k-1})$, $a_{k-1} = \tilde{f}(\hat{x}_{k-1\mid k-1};\mu) -F_j(\hat{x}_{k-1\mid k-1}) \hat{x}_{k-1\mid k-1}$, $H_k = H_j(\hat{x}_{k\mid k-1})$, $b_k = h(\hat{x}_{k \mid k-1}) - H_{j}(\hat{x}_{k\mid k-1}) \hat{x}_{k\mid k-1}$, $\Lambda_{k-1} = Q_{k-1}$,  and $\Omega_k = R_k$. Here, 
$F_j(\hat{x}_{k-1\mid k-1}) = \pdv{\tilde{f}(x_{k-1}; \mu)}{x_{k-1}}\Big|_{x_{k-1} = \hat{x}_{k-1\mid k-1}}$ and $H_{j}(\hat{x}_{k\mid k-1}) = \pdv{h(x_{k})}{x_{k}}\Big|_{x_{k} = \hat{x}_{k\mid k-1}}$ denote the Jacobians of the process and measurement models, respectively.

\subsection{Gaussian approximated Bayesian filter}

In this section, we present the approximate Bayesian filtering algorithm based on the affine approximation in \eqref{Eq_affine_approx}. The filtering procedure is performed in two steps \cite{sarkka2023bayesian}: prediction step and update step. On the prediction step, the predictive density of the state $x_k$ given the measurements $y_{1:k-1}$ is computed as $p(x_k \mid y_{1:k-1}) = \mathcal{N}(x_k \mid \hat{x}_{k\mid k-1}, P_{k\mid k-1})$, where
\begin{align}
   & \hat{x}_{k\mid k-1} = A_{k-1} \hat{x}_{k-1\mid k-1} + a_{k-1}, \label{prior_mean} \\&
    P_{k\mid k-1} = A_{k-1}P_{k-1\mid k-1}A_{k-1}^\top+ \Lambda_{k-1}, \label{prior_error_cov}
\end{align}
and the linearization parameters are computed as described in the previous section. Upon receiving the new measurement, the posterior distribution of $x_k$ conditioned on $y_{1:k}$ is given by $p(x_k \mid y_{1:k}) = \mathcal{N}(x_k\mid \hat{x}_{k\mid k},\, P_{k\mid k})$,
where
\begin{align}
    K_k & = P_{k\mid k-1} H_k^\top \left( P^{yy}_{k\mid k-1} \right)^{-1}, \label{Eq_K_gain}\\
    \hat{x}_{k\mid k} &= \hat{x}_{k\mid k-1} + K_k (y_k - \hat{y}_{k\mid k-1}), \label{Eq_post_mean} \\
      P_{k\mid k} &= P_{k\mid k-1} - K_k  P^{yy}_{k\mid k-1} K_k^{\top}. \label{Eq_post_cov}
\end{align} 
Above, the linearization parameters are again computed as described in the previous section,  $\hat{y}_{k\mid k-1} = H_k \hat{x}_{k\mid k-1} + b_k$ is the predicted mean of measurement, and $P^{yy}_{k\mid k-1} =  H_k P_{k\mid k-1} H_k^\top + \Omega_k $ is its error covariance.

The neural network parameters $\mu$ appearing in the SSM in \eqref{Eq_dynamic_model} are estimated by minimizing the negative log-likelihood, also known as the energy function, of the measurements $y_{1:T_s}$ given $\mu$ \cite{sarkka2023bayesian, kokkala2016sigma}. The energy function accumulated over $T_s$ time steps can be written as $ E_{T_s}(\mu) = \sum_{k=1}^{T_s} \frac{1}{2} \Big[ \log( (2 \pi)^{n_y} |P^{yy}_{k \mid k-1}|)   +  (y_k - \hat{y}_{k\mid k-1})^\top (P^{yy}_{k\mid k-1})^{-1} (y_k - \hat{y}_{k\mid k-1}) \Big].$
The implementation of the estimation algorithm, which also evaluates the energy function, is summarized in Algorithm \ref{Algo_DKF_affine}. 

\begin{algorithm}[h!]
		\caption{Estimation algorithm for affine SSM}\label{Algo_DKF_affine}
		\begin{algorithmic}[1]
  \Function{$[\{\hat{x}_{k|k}, P_{k|k}\}_{k=1}^{T_s}, E_{T_s}(\mu)] = \text{EST}$}{$y_{1:T_s}, \mu$}
 \State {\bfseries Initialize:} Start from $\hat{x}_{0|0}$, $P_{0|0}$, and $E_0(\mu) = 0$. 
  \For{$k = 1, \ldots, T_s$}
  \State Compute $[A_{k-1}, \, a_{k-1}, \, \Lambda_{k-1}]$ using SLR or Taylor \Statex \quad \, series approximation. 
  \State Compute $\hat{x}_{k\mid k-1}$ and $P_{k\mid k-1}$ following \eqref{prior_mean} and \eqref{prior_error_cov}. 
  \State Evaluate $[H_{k}, \, b_{k}, \, \Omega_{k}]$ using SLR or Taylor series \Statex \quad \, approximation. 
\State Update $\hat{x}_{k\mid k}$ and $P_{k\mid k}$ using \eqref{Eq_post_mean} and \eqref{Eq_post_cov}. 
\State Evaluate $E_k(\mu)=E_{k-1}(\mu)+\frac{1}{2}(\log (2 \pi)^{n_y} |P_{k\mid k-1}^{yy}|$ \Statex \quad\, $+(y_k-\hat{y}_{k\mid k-1})^\top(P_{k\mid k-1}^{yy})^{-1}(y_k- \hat{y}_{k\mid k-1}))$.
\EndFor
\EndFunction
  \end{algorithmic}
  \end{algorithm}

The parameter estimate is then obtained as \cite{sarkka2023bayesian}
\begin{equation}\label{Eq_min_energy}
    \hat{\mu} = \arg \min_{\mu} E_{T_s}(\mu),
\end{equation} 
which in practice can be found by using an iterative optimization method such as gradient descent or Adam optimizer \cite{goodfellow2016deep}.

The pseudo-code for parameter estimation is presented in Algorithm \ref{alg_training_EKF}. For simplicity, the update step in the algorithm is formulated using gradient descent \cite{goodfellow2016deep}. However, the proposed framework is general and can be implemented with other optimization methods, such as Adam.

\begin{algorithm}[tb]
   \caption{Parameter estimation for the proposed method}
   \label{alg_training_EKF}
\begin{algorithmic}[1]
   \State {\bfseries Initialize:} Start from an initial guess $\mu$. 
   \For{epoch $= 1, \ldots ,\textrm{MaxEpochs}$}
      \State Run the filter over the entire dataset to compute the \Statex \quad energy function $[\cdot, \, E_{T_s}(\mu)]=\text{EST}(y_{1:T_s}, \mu)$. 
      \State Compute the gradient of $ E_{T_s}$ with respect to 
      \Statex \quad $\mu$, that is, $ \nabla E_{T_s}(\mu)$. 
      \State Update the parameters $ \mu \gets \mu - \gamma \nabla E_{T_s}(\mu)$.
   \EndFor
\end{algorithmic}
\end{algorithm}
\section{Experimental Results}
In this section, we evaluate the proposed method on a simple pendulum and a Duffing oscillator. The kinetic energy is assumed symmetric in velocity, $T(\dot{q}) = T(-\dot{q})$, and we enforce this property by parameterizing it as a function of the squared velocity, using $\dot{q}^2$ as the input to the network. Additionally, we impose the condition $T(0) =0$ by subtracting the network output evaluated at zero velocity, thereby ensuring that the kinetic energy is zero when $\dot{q}=0$. The kinetic and potential energy networks are fully connected neural networks with one and two hidden layers of 32 and 64 neurons, respectively. Both use the softplus activation function, followed by a linear output layer without bias. To fix the reference level of the potential energy, we set $V(0) = 0$ by subtracting the network output at $q = 0$. Both networks are trained using the Adam optimizer with a learning rate of $5\times 10^{-3}$.  The dataset is divided into training and testing subsets, with 70\% of the data used for model training and the remaining 30\% reserved for testing and performance evaluation.

We implement the proposed EKF (PrEKF) and proposed cubature Kalman filter (PrCKF), and compare their performance with LNNs, the traditional EKF (TrEKF), and the traditional CKF (TrCKF), assuming the model is known. In the traditional LNNs, the velocity $\dot{q}$ is obtained via numerical differentiation, and $(q,\,\dot{q})$ are used as input to the LNNs.

For the simple pendulum problem, the states $(q,\, \dot{q})$ represent the angular position and angular velocity, respectively. The ground truth angular acceleration is given by $\ddot{q}=-\frac{g}{l}\sin{q}$ \cite{sarkka2023bayesian}, where $g$ is gravitational acceleration and $l$ is pendulum length. The mass and length of the pendulum are set to unity, and $Q_{k}$ is the same as in \cite[pp.~67--68]{sarkka2023bayesian}. The noisy measurement model is $y_k = q_k + \nu_k$, where $\nu_k \sim \mathcal{N}(0, \, 0.01)$.

The results summarized in Table \ref{tab:RMSE_noisydata} and Table \ref{tab:RMSE_withoutnoisedata} indicate that the proposed methods (PrEKF and PrCKF) achieve RMSE values comparable to those of standard estimators (TrEKF and TrCKF) with a known model for both noisy and noise-free measurements. On the other hand, the RMSE obtained using LNNs is significantly higher for noisy $y_k$ because the measurement noise in $q$ is amplified during numerical differentiation used to obtain $\dot{q}$. However, as reported in Table \ref{tab:RMSE_withoutnoisedata}, in the absence of measurement noise in $q$, the LNNs also provide results comparable to the traditional and proposed filters.

\begin{table}[]
    \centering
    \caption{RMSE values of different methods under noisy measurements for the simple pendulum and Duffing oscillator.}
    \label{tab:RMSE_noisydata}
    \begin{tabular}{|c|c|c|c|c|}\hline
    &\multicolumn{2}{c|}{\textbf{Simple pendulum}}&\multicolumn{2}{c|}{\textbf{Duffing oscillator}}\\\hline
        \textbf{Methods}& $q$ (rad) &$\dot{q}$ (rad/s)& $q$ (m) &$\dot{q}$ (m/s)\\\hline
        TrEKF &0.02 &0.07&0.02&0.04\\\hline
        TrCKF &0.02 &0.06&0.02&0.03\\\hline
        PrEKF&0.02 &0.16&0.03&0.08\\\hline
        PrCKF&0.02 &0.17&0.02&0.08\\\hline
        LNN&0.09&7.06&0.10&7.23\\\hline
    \end{tabular}
\end{table}

\begin{table}
    \centering
\caption{RMSE values of different methods for the simple pendulum and Duffing oscillator under noise-free $y_k$.}
\label{tab:RMSE_withoutnoisedata}
    \begin{tabular}{|c|c|c|c|c|}\hline
    &\multicolumn{2}{c|}{\textbf{Simple pendulum}}&\multicolumn{2}{c|}{\textbf{Duffing oscillator}}\\\hline
        \textbf{Methods}& $q$ (rad) &$\dot{q}$ (rad/s)& $q$ (m) &$\dot{q}$ (m/s)\\\hline
        TrEKF &0.01 &0.05&0.01&0.02\\\hline
        TrCKF &0.01 & 0.04&0.01 &0.03 \\\hline
        PrEKF& 0.01& 0.11& 0.01& 0.03\\\hline
        PrCKF& 0.02 & 0.14&0.01 &0.04 \\\hline
        LNN&0.00 &0.04&0.00&0.21\\\hline
    \end{tabular}
\end{table}

In the Duffing oscillator problem \cite{li2017state}, the states $(q,\,\dot{q})$ denote the position and velocity, respectively. The acceleration is given by $\ddot{q}=-\alpha q-\beta q^3$, where $\alpha=-1$ N/m and $\beta=1$ N/m$^3$, and $Q_k = 10^{-5} I_{2 \times 2}$. The measurement equation is $y_k = q_k + \nu_k$, where $\nu_k \sim \mathcal{N}(0, 0.01)$. The RMSE values reported in Table \ref{tab:RMSE_noisydata} and Table \ref{tab:RMSE_withoutnoisedata} show trends similar to those observed in the simple pendulum problem, with PrEKF and PrCKF achieving performance comparable to TrEKF and TrCKF with a known model, while the LNNs exhibit degraded performance for noisy measurements but comparable performance in the noise-free case.
\section{Conclusion}
In this paper, we have proposed a Bayesian filtering-based framework for learning system dynamics from partial, noisy measurements using a Lagrangian mechanics formulation. The kinetic and potential energies were parameterized using neural networks, and unknown forces were modeled as white Gaussian noise. A continuous-time stochastic state-space model was derived from the Lagrangian formulation. The maximum-likelihood approach was employed to jointly estimate the system states and neural network parameters using Gaussian approximation-based Bayesian filters. Numerical examples involving a simple pendulum and a Duffing oscillator demonstrated the effectiveness of the proposed approach and highlighted its advantages over conventional Lagrangian neural networks.

\bibliographystyle{IEEEtran}
\bibliography{Bibliography}

\end{document}